\pdfoutput=1

\documentclass[11pt]{article}

\usepackage[preprint]{acl}

\usepackage{times}
\usepackage{latexsym}

\usepackage[T1]{fontenc}

\usepackage[utf8]{inputenc}

\usepackage{microtype}

\usepackage{inconsolata}

\usepackage{graphicx}

\usepackage{lipsum}
\usepackage{algorithm}
\usepackage{algorithmic}
\usepackage{amsmath}
\usepackage{soul}
\usepackage{array} 

\title{DAMRO: Dive into the Attention Mechanism of LVLM to Reduce Object Hallucination}

\author{Xuan Gong, Tianshi Ming, Xinpeng Wang, Zhihua Wei\thanks{Corresponding author} \\
        Department of Computer Science and Technology, Tongji University\\
        \texttt{ \{2152095,2151569,wangxinpeng,zhihua\_wei\}@tongji.edu.cn}}

\begin{document}
\maketitle
\begin{abstract}

 Despite the great success of Large Vision-Language Models (LVLMs), they inevitably suffer from hallucination. As we know, both the visual encoder and the Large Language Model (LLM) decoder in LVLMs are Transformer-based, allowing the model to extract visual information and generate text outputs via attention mechanisms. 
 We find that the attention distribution of LLM decoder on image tokens is highly consistent with the visual encoder and both 
 distributions tend to focus on particular background tokens rather than the referred objects in the image. We attribute to the unexpected attention distribution to an inherent flaw in the visual encoder itself, which misguides LLMs to over emphasize the redundant information and generate object hallucination. To address the issue, we propose DAMRO, a novel training-free strategy that \textbf{D}ive into \textbf{A}ttention \textbf{M}echanism of LVLM to \textbf{R}educe \textbf{O}bject Hallucination. Specifically, our approach employs classification token (CLS) of ViT to filter out high-attention outlier tokens scattered in the background and then eliminate their influence during decoding stage. We evaluate our method on LVLMs including LLaVA-1.5, LLaVA-NeXT and InstructBLIP, using various benchmarks such as POPE, CHAIR, MME and GPT-4V Aided Evaluation. The results demonstrate that our approach significantly reduces the impact of these outlier tokens, thus effectively alleviating the hallucination of LVLMs. The code is released at \url{https://github.com/coder-gx/DAMRO}.
\end{abstract}

\section{Introduction}

Large Vision-Language Models (LVLMs) research \citep{dai2023instructblip, liu2024llavanext,chen2023minigptv2,Ye2023mPLUGOwI2RM} has witnessed rapid advancement in the past few years, particularly demonstrating strong capabilities in visual reasoning tasks. However, LVLMs still face significant challenges related to object hallucination \citep{objectHallucination}, where the objects described in the generated text do not align with the visual ground truth of the input. This issue is prevalent across various models, posing a critical problem for the reliability and safety of LVLMs \citep{ahmad2023creating}.


Recently, the issue of object hallucination in LVLMs has gained increasing attention. Early work has tried many methods, such as optimizing the training and fine-tuning methods \citep{sarkar2024mitigating, xiao2024detecting}, incorporating external information or models, e.g. DETR \citep{10.1007/978-3-030-58452-8_13}\citep{zhao2024mitigating,chen2024halc}, providing feedback on hallucinated information and reprocesses \citep{zhou2024analyzing,yin2023woodpecker}. Efforts also include LLM decoding methods, like contrastive decoding \citep{Leng_2024_CVPR,Favero_2024_CVPR} and other novel decoding methods \citep{Huang_2024_CVPR}.

These approaches mainly focus on improving the overall model architecture or specific modules within LVLMs, such as the visual encoder or LLM decoder. However, they often overlook the fundamental component of LVLMs, the Vision Transformer (ViT) structure \citep{dosovitskiy2021an}, and its impact on the hallucination generation mechanism during the LLM decoding stage.


Based on LLaVA-1.5 \citep{Liu_2024_CVPR}, we explore the attention map in both the visual encoder and the LLM decoder. We find outlier tokens in the attention map of both components, which are highly consistent with each other. These high-norm outlier tokens often contain globally redundant visual information \citep{darcet2024vision}. Additionally, our analysis reveals a correlation between attention to these tokens and the occurrence of object hallucination.



\begin{figure*}[t]
  \includegraphics[width=1.0\linewidth]{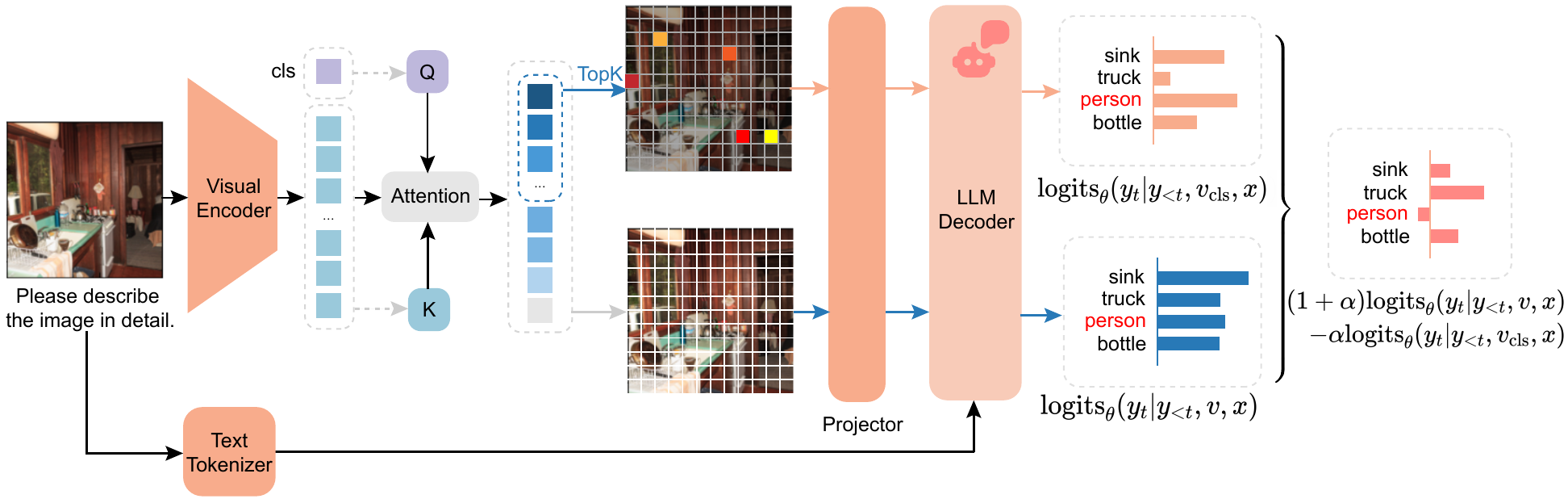} \hfill
  \caption {An overview of DAMRO. We utilize attention mechanism to filter the outlier tokens, and then apply contrastive decoding to mitigate the influence of outlier tokens in LLM decoding stage.}
  \label{fig:overview}
\end{figure*}

To address the aforementioned issue, we propose the \textbf{D}ive into \textbf{A}ttention \textbf{M}echanism of LVLM to \textbf{R}educe \textbf{O}bject Hallucination (DAMRO) method, as illustrated in Figure~\ref{fig:overview}. DAMRO filters out high-norm outlier tokens from the ViT attention map, identifying them as negative tokens, and then projects them into the LLM along with normal tokens. Contrastive decoding is then applied to reduce the LLM decoder's reliance on these tokens that contain globally redundant information and to enhance its focus on object-level details, thus mitigating model hallucination.

Our method is training-free and does not introduce external information or models. It outperforms similar approaches such as M3ID \citep{Favero_2024_CVPR} and VCD \citep{Leng_2024_CVPR} in overall effectiveness. Additionally, since ViT is such a popular backbone of visual encoder \citep{yin2024survey} that our approach demonstrates strong generalizability due to its utilizing on attention mechanism.


In conclusion, our main contributions are summarized as follows:

\begin{itemize}
    \item We conduct in-depth analysis of the relationship between the attention maps of the visual encoder and the LLM decoder, revealing a high consistency in the distribution of their outlier tokens.
    \item We analyze the impact of the consistency on object hallucination and design the DAMRO method to mitigate the hallucination in LVLMs.
    \item We demonstrate effectiveness of our method via extensive experiments on various models and benchmarks. Moreover, our training-free approach is applicable to most LVLMs without external knowledge or models.
\end{itemize}

\section{Related Work}
\subsection{Hallucination in LVLMs}

In LVLMs, hallucination refers to discrepancies between visual input (ground truth) and textual output. Hallucination is initially identified and studied in LLM research~\citep{huang2023survey,Ji_2023}. However, LVLMs also suffer from hallucination, which is much more complex due to their intricate structure. \citet{han2024instinctive} analyze hallucination from the perspective of training data bias. \citet{Tong_2024_CVPR}, \citet{Jiang_2024_CVPR}, and \citet{Huang_2024_CVPR} focus on structural causes, revealing the flaws in visual encoders, the misalignment of visual-textual modalities, and the inherent hallucinations of LLM respectively. \citet{zhou2024analyzing} identify patterns in LVLM input and output, proposing object co-occurrence, model uncertainty, and the spatial positioning in sentence as causes. These studies reveal the mechanisms of hallucinations and offer new approaches to address this issue in LVLMs.


Unlike previous studies, we start by analyzing the attention maps of the visual encoder and LLM decoder, focusing on their distribution characteristics and correlations. This analysis provides new insights into object hallucination.

\begin{figure*}[t]
  \includegraphics[width=\linewidth]{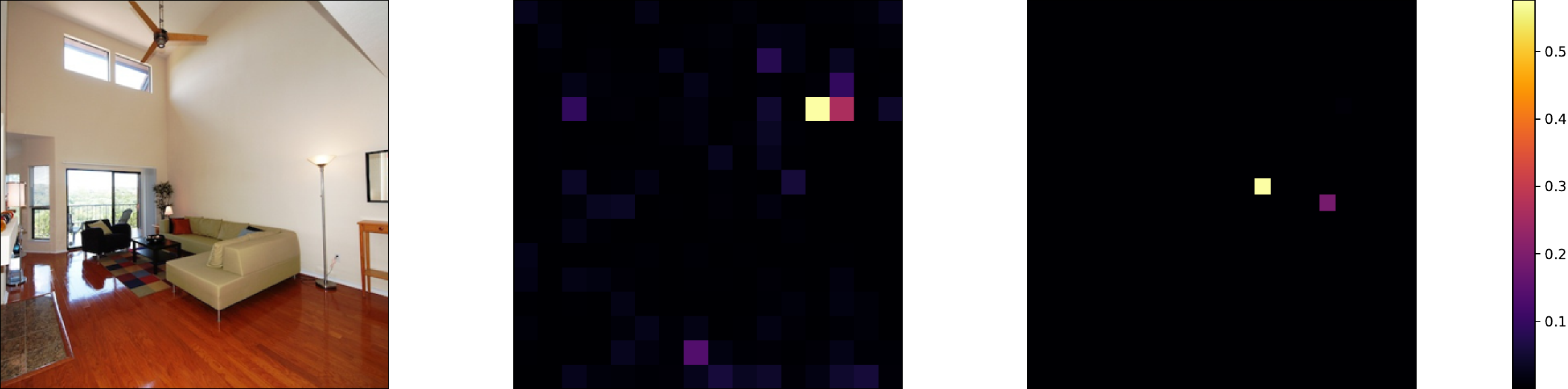}
  \caption{Attention map of visual encoder. \textbf{Left:} original image. \textbf{Middle:} attention map of InstructBLIP ViT (16x16). \textbf{Right:} attention map of LLaVA-1.5 ViT (24x24).}
  \label{fig:cls}
\end{figure*}

\subsection{Contrastive Decoding to Mitigate Hallucination}

Contrastive decoding \citep{li-etal-2023-contrastive} is first introduced in text generation tasks in LLMs to reduce noise by subtracting the distribution of an amateur model. To address hallucination issues in LVLMs, researchers have introduced contrastive decoding to improve model performance. \citet{Leng_2024_CVPR} apply Gaussian noise to images to increase visual uncertainty. They use these noisy images as negative samples to subtract the LLM's prior and reduce object hallucination. \citet{Favero_2024_CVPR} employ pure text inputs as negative samples. They apply contrastive decoding to enhance the influence of visual information during text generation. \citet{wang-etal-2024-mitigating} introduce a disturbance instruction to force the model to output an error distribution, which is then subtracted to mitigate hallucination. 

Given that our method draws on contrastive decoding 
and considering the generality and effectiveness of these methods, in section \ref{sec:baseline}, we select VCD \citep{Leng_2024_CVPR} and M3ID \citep{Favero_2024_CVPR} as our baselines for experimental comparison.

\section{Motivation}
\subsection{Problem Formulation}

We segment the LVLM generation process into three distinct stages: Visual Encoding, Projection, and LLM Decoding. In the initial stage, an input image is divided into \(n\) patches, each projected into a token embedding via Vision Transformer. The set of \(n\) tokens is represented as \(X_v=\{X_{v_i} | 0 \leq i < n\}\) . Then tokens are forwarded to the LLM after projection. Concurrently, the prompt is tokenized into tokens \(X_l\) and is put into the LLM directly or indirectly.





In the decoding stage, we perform autoregressive decoding with the transformer, which is formulated in Eq.~\ref{eq:e3}.
\begin{equation}
  \label{eq:e3}
      p_t= \text{softmax}(\text{logits}_{\theta}(y_t|y_{<t},X_v,X_l)).  
\end{equation}
where $p_t$ represents  probability distribution  of next token $y_t$ in the $t$-th step of decoding, $y_{<t}$  represents the generated  text from $0$ to $t-1$ step and $\text{logits}_{\theta}$ represents the logit distribution. Then the LLM adopts a specific strategy to obtain the next token based on the probability distribution $p_t$.

We studied the impact of the visual token $X_v$ on \(\text{logits}_{\theta}(y_t|y_{<t},X_v,X_l)\) to reduce the likelihood of hallucination  occurrence.

\subsection{Drawbacks of ViT}\label{sec:vit_bad}

The Vision Transformer \citep{dosovitskiy2021an} has gained widespread favor as the backbone visual encoder for all LVLMs due to its superior visual representation capabilities. However, \citet{darcet2024vision} find that there are always high-norm outlier tokens in ViT, which tend to appear in background regions with redundant patch information, containing minimal local information but a little global information.






The attention map of LVLMs' visual encoder also focus on a small number of high-norm outlier tokens, as illustrated in Figure~\ref{fig:cls}. We posit that these outlier tokens embody the negative visual priors within the ViT. And when image tokens are projected and sent to the LLM, the LLM also tends to focus on these tokens due to their high attention value in visual encoder, leading to the ignorance of local information contained within other patches. This may result in a degradation of the model's fine-grained visual capabilities.


 To validate the information contained within these tokens as perceived by the LLM, we conducted ablation experiments (results provided in Appendix~\ref{sec:token}). The findings confirmed that these few tokens indeed contain substantial information, but are not accurate enough.

\subsection{Outlier Tokens Cause Hallucination}\label{sec:ana}

\begin{figure}[t]
  \includegraphics[width=1.0\linewidth]{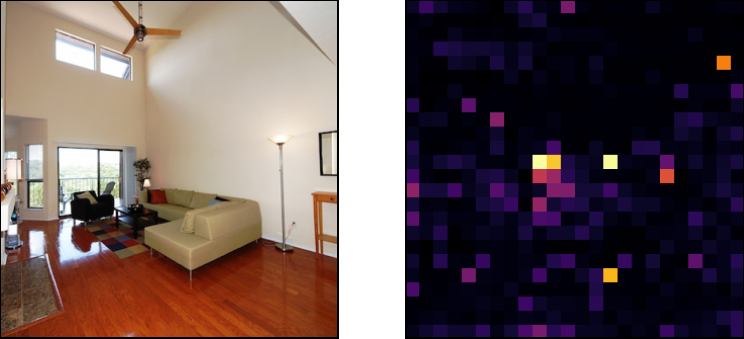}
  \caption {LLM decoder attention map of "plant" token (non-hallucinatory). It is evident that attention can accurately locate the position of the plotted plant.}
  \label{fig:non-ha-map}
\end{figure}
\begin{figure}[t]
  \includegraphics[width=\linewidth]{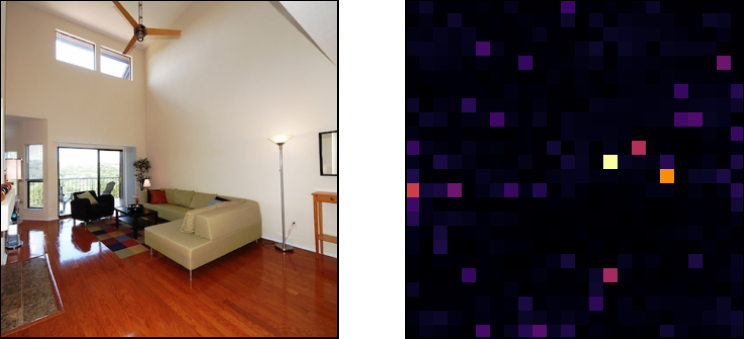}
  \caption {LLM decoder attention map of "clock" token (hallucinatory). The attention mainly focus on the outlier tokens in the background, whose positions are the same in visual encoder attention map in the right sub-image of Figure~\ref{fig:cls}.}
  \label{fig:ha-map}
\end{figure}
\begin{figure}[t]
  \includegraphics[width=\linewidth]{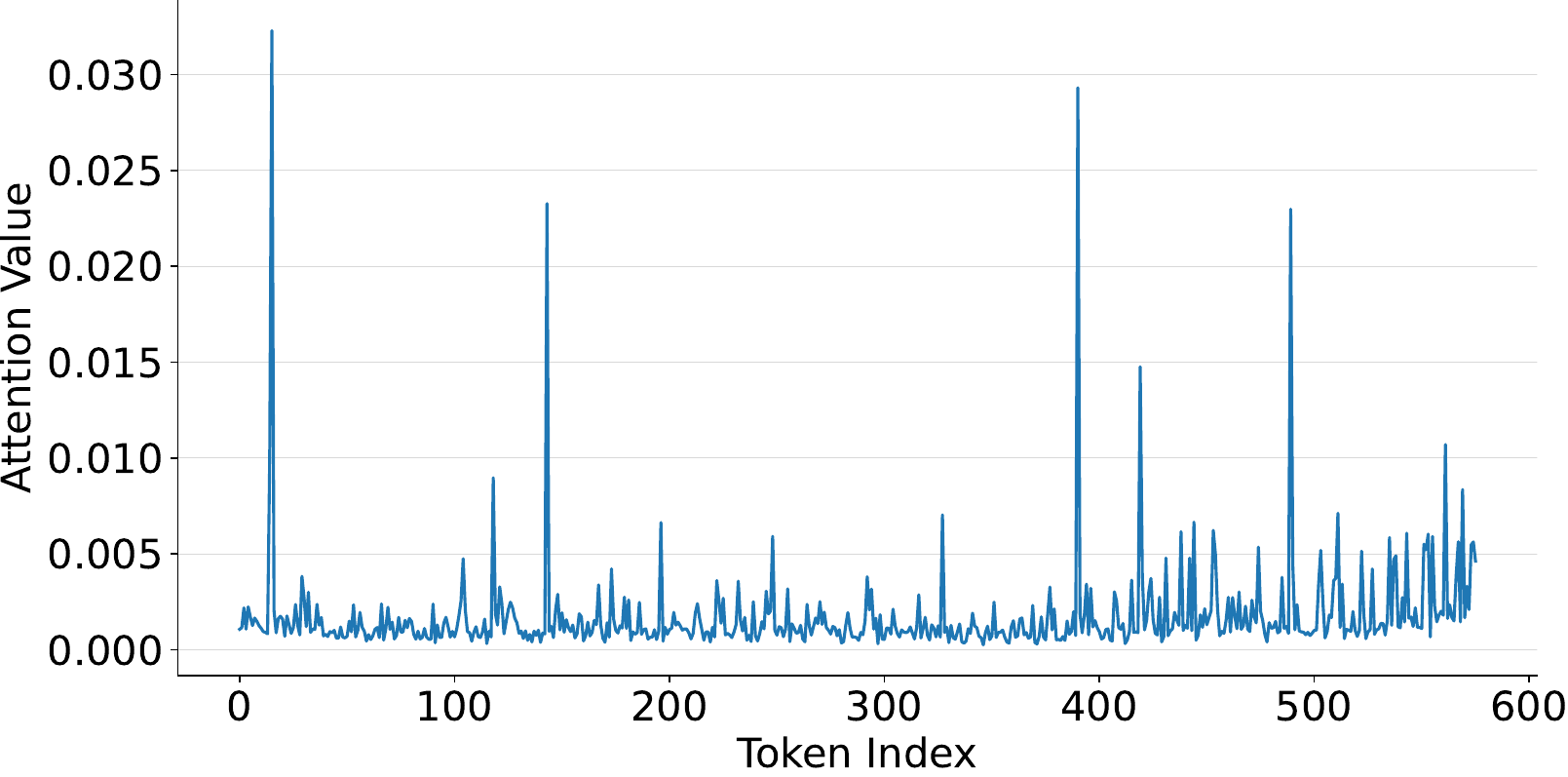}
  \caption {The proportion of the overall attention map in LLM decoder.}
  \label{fig:LLM_att}
\end{figure}
\begin{figure}[t]
  \includegraphics[width=1\linewidth]{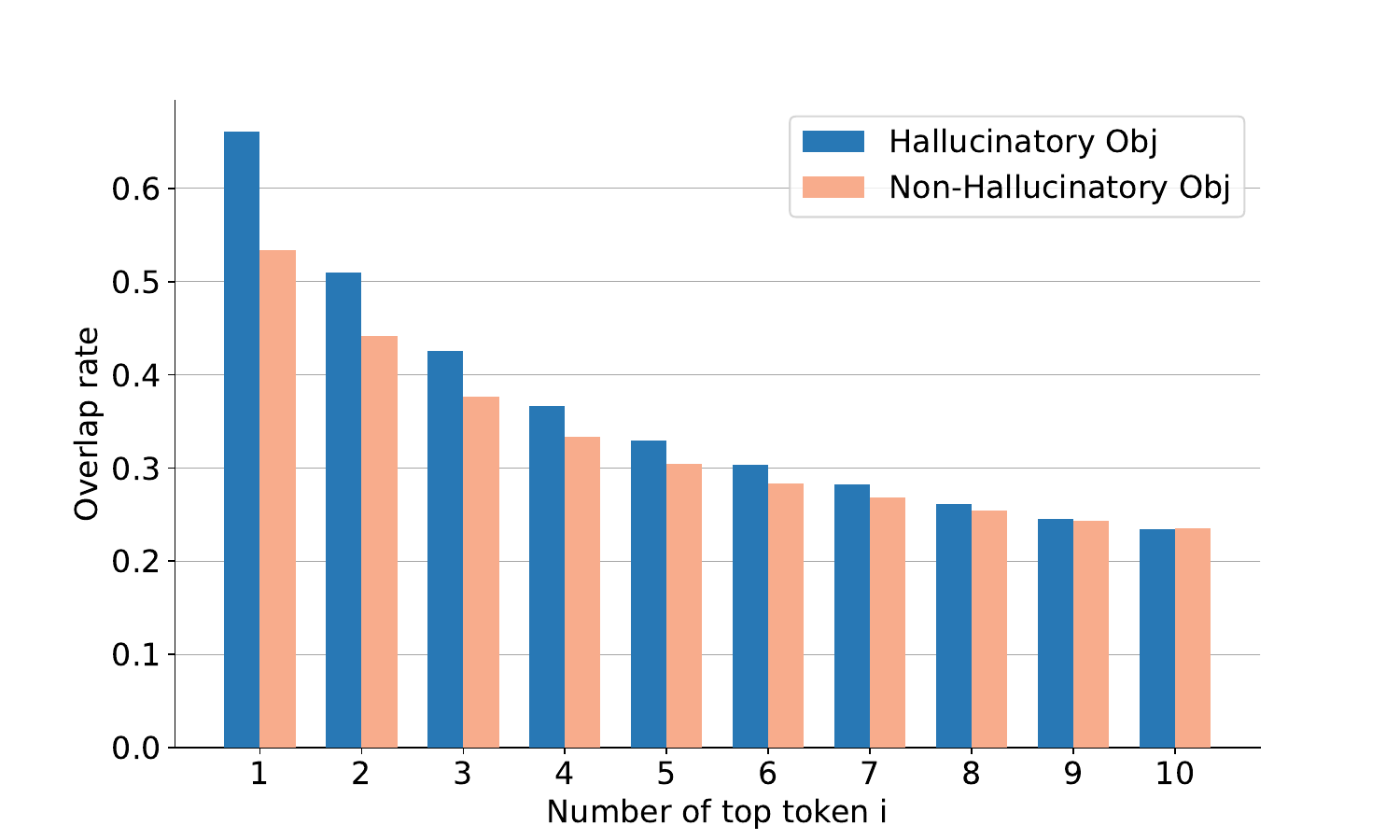}
  \hfill
  \includegraphics[width=1\linewidth]{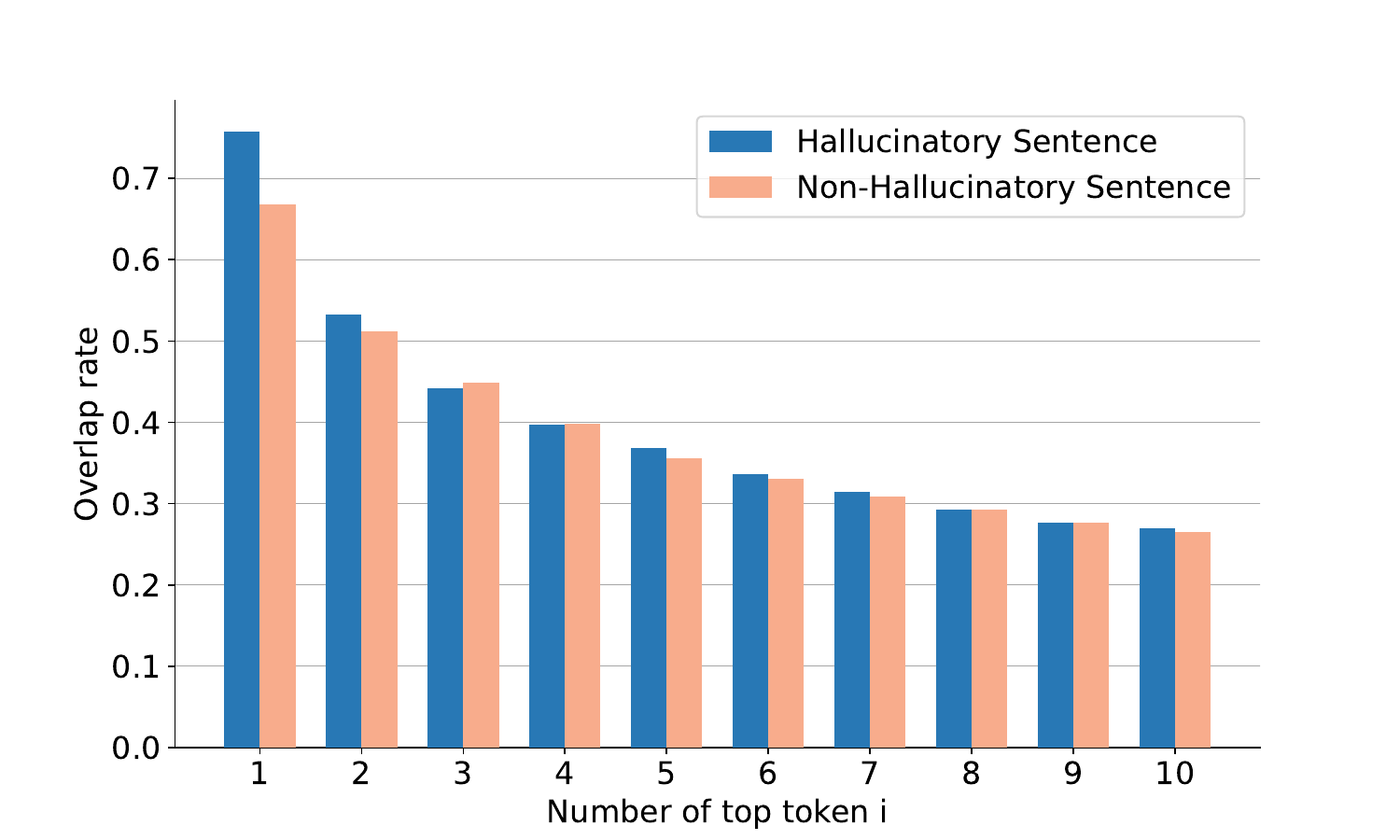}
  \caption {Top 1-10 outlier tokens overlap rate between visual encoder and LLM decoder. Both of object-level and sentence-level results show that hallucination tends to happen when overlap rate is higher, especially considering the top tokens.}
  \label{fig:overlap}
\end{figure}

Based on the aforementioned issues in ViT, we attempt to observe the attention maps of image tokens during LLM decoding stage. We find that LLM decoder attention map also features with a few outlier tokens at the same position as visual encoder that get most of the attention compared to other tokens, as illustrated in Figure~\ref{fig:LLM_att}. We assume that this consistency is related to the occurrence of hallucination, where the LLM decoder pays more attention to outlier tokens identified in visual encoding stage. And we selected an example (Figure~\ref{fig:non-ha-map},~\ref{fig:ha-map}) to demonstrate this correlation. To quantitatively characterize the consistency, we propose an evaluation metric \(H_i\),  where \(S_v(i)\) denotes the set of top \(i\) tokens of attention value from the visual encoder's attention map, while \(S_l(i)\) represents the set of top \(i\) tokens from the LLM decoder's attention map. And in this formulation, \(|S|\) denotes the cardinality of the set \(S\), which is the number of elements contained within \(S\).
\begin{equation}
  \label{eq:e4}
   H_i= \frac{|S_{v}(i)\cap S_{l}(i)| }{i}.
\end{equation}
We randomly select 1000 images from the val2014 subset in MSCOCO dataset~\citep{10.1007/978-3-319-10602-1_48} and query LLaVA-1.5 with the prompt "What can you see in this image ?" to get the descriptions from model. We use the generated captions and object words as two kinds of units and employed CHAIR~\citep{objectHallucination} to identify hallucinations. We then utilize metric \(H_i\) to analyze the relation between the occurrence of hallucinations and the consistency of their distributions, as illustrated in Figure~\ref{fig:overlap}.

\begin{figure}[t]
  \includegraphics[width=\linewidth]{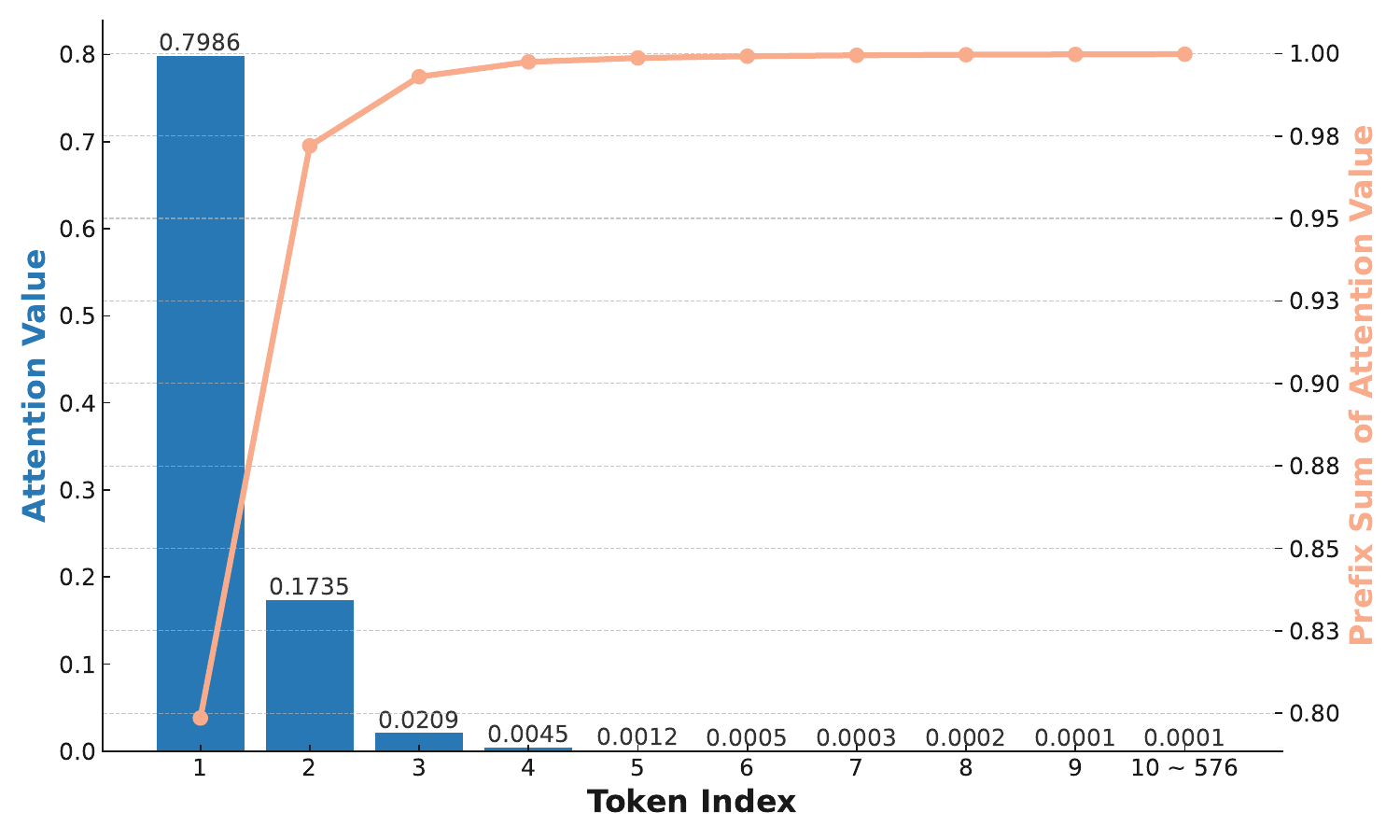}
  \caption {The proportion of the overall attention map occupied by tokens sorted by attention value in visual encoder.}
  \label{fig:topcls}
\end{figure}

Additionally, we found that the top three tokens with the highest attention score in the visual encoding stage accounted for more than 99\% of the attention, as shown in Figure~\ref{fig:topcls}. To further verify the influence of these tokens, we analyzed the proportion of the same three tokens\footnote{Unless otherwise specified, in this paper, the same tokens in the visual encoder and LLM decoder refer to tokens corresponding to the same spatial positions in the image.} in the attention map of LLM decoder. The evaluation metric of the influence is denoted as \(F\), defined as
 \begin{equation}
  \label{eq:e5}
   F=  \frac{\sum^3_{j=1}ATT(L_v(j))}{\sum_{i=0}^{n-1} ATT(i)}.
\end{equation}
where \(L_v(i)\)  represents the position of the token with \(i\)-th highest attention value in the visual encoder attention map and \(ATT(i)\) represents the LLM decoder attention value of the token at position \(i\).


Similarly, we use generated captions and object words as units to identify hallucinations. And we get the $F$ results in Table~\ref{tab:influence}. It can be observed that outlier tokens in visual encoding stage indeed have influence on the subsequent LLM decoding stage, which is closely related to the occurrence of hallucinations. 

\begin{table}
  \centering
  \begin{tabular}{c c c}
    \hline
        \textbf{Granularity} & \textbf{HA} &  \textbf{Non-HA} \\
    \hline
         {sentence-level}    & {0.0554}      & {0.0539}  \\
          {object-level}    & {0.0605}         & {0.0551} \\
    \hline
  \end{tabular}
  \caption{$F$ Value results. HA: hallucinatory, Non-HA: non-hallucinatory. It is easily observed that at both the sentence level and the object level, the influence of outlier tokens from the visual encoder is greater when hallucinations occur.}
  \label{tab:influence}
\end{table}


\section{Methods}

\subsection{Outlier Tokens Selection}

In the final layer of self-attention in ViT, the class token [CLS] is generally used for classification \citep{dosovitskiy2021an}. The [CLS] token is used as the query vector in attention calculation with other visual tokens as key vector:
\begin{equation}
  \label{eq:e6}
   A_{\text{cls}}= \text{softmax}\left(\frac{Q_{\text{cls}}K^{T}}{\sqrt{d}}\right).
\end{equation}
where \(Q_{cls}\) is the result of the [CLS] token's query vector after being multiplied by the corresponding weights; \(K^T\) is the result of all other image tokens' key vectors after being multiplied by their corresponding weights, and \(d\) is the dimension of \(Q_{cls}\).

We sample the top \(k\) outlier tokens based on attention value between the class token [CLS] and spatial visual tokens at the second-to-last layer, which is denoted as:
\begin{equation}
  \label{eq:e7}
\text{token}_{\text{outlier}}=\mathop{\arg\max}\limits_{\text{token}_i}(A_{\text{cls}}(\text{token}_i)).
\end{equation}
For the selection of the top \( k \), it is important to note that LLaVA-1.5 \citep{Liu_2024_CVPR} and InstructBLIP \citep{dai2023instructblip} have different ViT structures. ViT in LLaVA-1.5 contains 576 (24x24) image tokens, whereas InstructBLIP has only 256 (16x16). The different numbers of image tokens lead to different choices in values of \( k \) for the top \( k \) selection. The difference in $k$ value will be discussed in detail in the ablation experiment in Appendix~\ref{sec:abla}.

\subsection{Contrastive Decoding}



We use Contrastive Decoding \citep{li-etal-2023-contrastive} to mitigate the impact of visual outlier tokens from the visual encoder on subsequent text generation. In LVLMs, Contrastive Decoding is typically conducted during the sampling process of LLM decoding, where the next token is determined based on the probability distribution in the logits space.

Answer generation in LLMs is an autoregressive process, in which the contrastive decoding is formulated as Eq.~\ref{eq:e8}.
\begin{equation}
  \label{eq:e8}
  \begin{split}
      p_t= \text{softmax} ((1+\alpha)\text{logits}_{\theta}(y_t|y_{<t},v,x) &\\ - \alpha \text{logits}_{\theta}(y_t|y_{<t},v_{\text{cls}},x)).&
  \end{split}
\end{equation}
where the probability distribution of the next token at step \( t \) is \( p_t \) with $x$ being the prompt input.  \( v_{\text{cls}} \in  v \) is visual information filtered by [CLS] token from overall visual information $v$.

The probability distribution in the logits space attenuates the influence of previous outlier tokens on decoding. This allows the model to focus more on fine-grained semantic information and eliminates redundant information containing visual encoder priors, thus mitigating hallucinations in the LVLM.

 To address the issue of excessive removal of global information, we introduced an adaptive plausibility constraint~\citep{li-etal-2023-contrastive}. In constrative decoding stage, we set a threshold \(\beta\) to truncate the new probability distribution based on the confidence level of the original model's predictions. The specific form is shown in Eq.~\ref{eq:e9}:
\begin{equation}
  \label{eq:e9}
   \begin{split}
       \mathcal{V}_{\text{head}}(y_{<t}) = &\{y_t \in    \mathcal{V} : p_{\theta}(y_t | v, x, y_{<t}) \\ &\geq \beta \max\limits_{w} p_{\theta}(w|v, x, y_{<t})\}.
   \end{split}
\end{equation}
\( \mathcal{V}_{\text{head}}\) serves as a filtering constraint for sampling the next token. The whole algorithm is further explained in Algo.~\ref{alg:all}.

\begin{algorithm}
    \caption{DAMRO}
    \label{alg:all}
    \begin{algorithmic}[1]
        \REQUIRE  
        text query $x$,
        image input $v$, 
        visual encoder $I_{\phi}$.
        \STATE Initialize empty output $y$ = [].
        \STATE Large Language Model $\mathcal{M}_{\theta}$.
        \FOR {t=0,1,2...}
            \STATE ${I_{\phi}(v)}_{i=1}^n \gets \text{VisualEncoder}(v)$
            \STATE $\log p_{\text{origin}}\gets\text{logits}_{\theta}(y_t|y_{<t},{I_{\phi}(v)}_{i=1}^n,x)$
            \STATE $\text{Attn}_{c}^{i}\gets\text{Attention}(\text{token}_{cls}, {I_{\phi}(v)}_{i=1}^n)$
            \STATE $I_{\text{outlier}} = \mathop{\arg\max}_{I}(\text{Attn}_{c}^{i})$
            \STATE $\log p_{\text{negetive}}\gets\text{logits}_{\theta}(y_t|y_{<t},I_{\text{outlier}},x)$
            \STATE Get token distribution in constrastive learning, $p_t\gets\text{softmax}((1+\alpha)\log p_{\text{origin}}-\alpha \log p_{\text{negetive}})$,
            \STATE Considering adaptive plausibility constraint, $p_t=p_t \text{ if } p_t \geq \max(\log p_{\text{origin}}) \text{ else } 0 $
            \STATE Get next token using random sample strategy $y_t$.
            \STATE $y=[y,y_t]$
            \IF{$y_t = \text{<EOS>}$}
                \STATE break
            \ENDIF
        \ENDFOR
        \RETURN Generated prompt $y$.
       
    \end{algorithmic}
\end{algorithm}

\section{Experiments}
\subsection{Experimental Settings}
\paragraph{LVLM Models}

We select three of the most representative LVLM models for evaluation: LLaVA-1.5-7b, LLaVA-NeXT-7b, and InstructBLIP-7b. For visual encoder, LLaVA-1.5 and LLaVA-NeXT share the same ViT backbone, both using ViT-L-336px pretrained from CLIP-L/14-336px \citep{radford2021learning}. In contrast, InstructBLIP uses ViT-g/14 pretrained from EVA-CLIP \citep{sun2023evaclip}. All three models use Vicuna\footnote{Vicuna-7b v1.5 for LLaVA-1.5 and LLaVA-NeXT, Vicuna-7b v1.1 for InstrutBLIP} \citep{chiang2023vicuna} as the LLM module.

Regarding the connection module between the two modalities, LLaVA-1.5 and LLaVA-NeXT use MLP layers to bridge feature gap between vision and text modalities without changing the amount of image tokens in the LLM. Conversely, InstructBLIP employs Q-Former \citep{10384565} for modality alignment, which standardized the number of visual tokens in LLM to 32.

Our approach is based on LLaVA-1.5 in the analysis of Section~\ref{sec:ana}. For more insights into generalizability, we also test our method on InstructBLIP, which has a significantly different structure compared to LLaVA-1.5, and we find that the performance still surpasses that of original model. This demonstrates that mitigating the impact of outlier tokens in the visual encoder is effective in alleviating hallucination across different projection modules.

\begin{table*}[]
\centering
\begin{tabular}{cccccc}
\hline

\textbf{Base Model} & \textbf{Method} & \textbf{Precision} & \textbf{Recall} & \textbf{F1 Score}    & \textbf{Accuracy} \\ \hline
LLaVA-1.5       & Original        & 88.63              & 73.76           & 80.48          & 82.08             \\
               & VCD             & 86.15              & \textbf{83.78}  & \textbf{84.87} & {\ul{84.98}}       \\
               & M3ID            & \textbf{92.48}     & 75.22           & 82.93          & 82.93             \\
               & \textbf{DAMRO}  & {\ul{88.84}}        & {\ul{81.09}}     & {\ul{84.72}}    & \textbf{85.31}    \\     \noalign{\vskip 0.8ex}  
               \hline
            
LLaVA-NeXT     & Original        & {\ul{92.28}}        & 75.58           & 83.07          & 84.57             \\
               & VCD             & 91.90              & \ul{82.4}      & {\ul{86.86}}    & {\ul{87.50}}       \\
               & M3ID            & \textbf{94.23}     & 79.2            & 86.05          & 80.87             \\
               & \textbf{DAMRO}  & 90.02              & \textbf{85.40}  & \textbf{87.60} & \textbf{87.87}    \\ \noalign{\vskip 0.8ex}   
               \hline
InstructBLIP   & Original        & 78.64              & 79.42           & 78.99          & 78.85             \\
               & VCD             & {\ul{84.88}}        & {\ul{79.93}}     & {\ul{81.96}}    & \textbf{82.56}    \\
               & M3ID            & \textbf{90.59}     & 70.58           & 79.33          & 81.60             \\
               & \textbf{DAMRO}  & 80.67              & \textbf{83.89}  & \textbf{82.20} & {\ul{81.77}}       \\ \noalign{\vskip 0.8ex}  
               \hline
\end{tabular}
\caption{Results of POPE. (The foundation model without methods is denoted as Original). The best value in the table is highlighted in \textbf{bold}, and the second best value is \ul{underlined}.}
\label{tab:pope}
\end{table*}

\paragraph{Baselines} \label{sec:baseline}

 We select two popular and training-free contrastive decoding methods: VCD \citep{Leng_2024_CVPR} and M3ID \citep{Favero_2024_CVPR}. Both approaches aim to enhance the impact of visual features during the LLM decoding phase by eliminating language priors. VCD generates negative logits using Gaussian blurring, while M3ID generates negative logits using pure text that without visual information. Additionally, we include the original model for comparison to highlight the improvements over the baseline model. For detailed experimental hyperparameter settings of these baselines, please refer to Appendix~\ref{sec:impl}.

\paragraph{Implementation Details}

Considering the characteristics of different visual encoders, for LLaVA-1.5 and LLaVA-NeXT, we set \(\alpha\) (Eq.~\ref{eq:e8}) to 0.5 for CHAIR benchmark and 2 for other benchmarks and we select top 10 (Eq.~\ref{eq:e7}) tokens as outlier tokens. For InstructBLIP, we set \(\alpha\) to 1.5 for CHAIR benchmark and 0.5 for other benchmarks and we select top 4 tokens as outlier tokens. To avoid introducing additional factors, we directly use the probability distribution generated by the softmax function as the sampling distribution and employ the basic random sampling decoding strategy. For all experiments, the seed is set to 42, max\_new\_token is set to 1024 and \(\beta\) (Eq.~\ref{eq:e9}) is set to 0.1 .

\subsection{Benchmarks and Experimental Results}\label{sec:exp}
\paragraph{POPE}


The Polling-based Object Probing Evaluation (POPE) \citep{li-etal-2023-evaluating} is a streamlined approach to assess object hallucination. LVLMs are required to respond to formatted questions in the form: "Is there a <object> in the image?" with "Yes" or "No," . The answers to these questions alternate between "Yes" and "No," ensuring an equal 50\% probability for each response. The complete POPE test is divided into three splits: random, popular and adversarial, in which missing objects are randomly selected, most frequently occurring in the dataset, and highly correlated with those present in the image respectively.

The dataset consists of 500 randomly selected images from the MSCOCO \citep{10.1007/978-3-319-10602-1_48} validation set. To facilitate testing, we add the prompt "Please use one word to answer this question." to restrict LVLM responses to "Yes" or "No". Four key evaluation metrics are generated: Precision, Recall, F1 score, and Accuracy. We average the results across the three splits, and the outcomes are presented in Table \ref{tab:pope}. More details are shown in Appendix~\ref{sec:de_pope}.


\begin{table}[]
\centering
\begin{tabular}{cccc}
\hline
\textbf{Model} & \textbf{Method} & \textbf{$\text{C}_S \downarrow$} & \textbf{$\text{C}_I \downarrow$} \\ \hline
LLaVA-     & Original        & 12.4              & 7.2                          \\1.5 
            & VCD             &  7.6               &  4.1                      \\
               & M3ID            & 9.2               & 5.3                         \\
               & \textbf{DAMRO }          &\textbf{ 6.0  }             & \textbf{ 3.6 }                      \\ 
                \noalign{\vskip 0.8ex}  
               \hline
LLaVA-        & Original        & 4.2               & 9.0                       \\NeXT
              & VCD             & \textbf{ 3.0 }              &\textbf{4.1}                      \\
               & M3ID            & 4.2               & 6.8                   \\
               &\textbf{ DAMRO}           & \textbf{3.0}               & 5.2              \\ 
                \noalign{\vskip 0.8ex}  
               \hline
Instruct-    & Original        & 7.8               & 5.2                        \\BLIP
              & VCD             & 3.2               & 1.9                        \\
               & M3ID            & 5.2               & 3.7                        \\
               & \textbf{DAMRO  }         & \textbf{ 2.8 }              & \textbf{1.7}                   \\   
               \hline
\end{tabular}
\caption{Results of CHAIR. $\text{C}_S$: $\text{CHAIR}_S$, $\text{C}_I$: $\text{CHAIR}_I$.}
\label{tab:chair}
\end{table}


\paragraph{CHAIR}


The Caption Hallucination Assessment with Image Relevance (CHAIR) \citep{objectHallucination} is a widely used metric for evaluating object hallucination in image captioning tasks. CHAIR compares the captions generated by the LVLM with the ground truth to identify correctly and incorrectly described objects in the captions. It then calculates the proportion of objects mentioned in the captions that are not present in the images CHAIR evaluates hallucination on two dimensions: $\text{CHAIR}_S$ and $\text{CHAIR}_I$. The former calculates the proportion of sentences containing hallucinations at the sentence level, while the latter computes the proportion of hallucinated objects out of all identified objects at the object level. These two metrics can be formulated as follows:

\begin{equation}
  \label{eq:e10}
  \begin{split}
      &\text{CHAIR}_{S}=\frac{|\{\text{captions w/ hallucinated objects}\}|}{|\{\text{all captions}\}|}. \\&\text{CHAIR}_{I}=\frac{|\{\text{hallucinated objects}\}|}{|\{\text{all mentioned objects}\}|}.
  \end{split}
\end{equation}


Similarly, we conducted the CHAIR evaluation on the MSCOCO dataset with 80 annotated object categories. We randomly selected 500 images from the validation set of COCO 2014 and used the prompt "Generate a short caption of this image." to obtain the generated captions.


The test results are shown in Table \ref{tab:chair}. It can be observed that, CHAIR scores on LLaVA-1.5 and InstructBLIP both surpassed the baseline compared to other methods, which achieve significant improvements in comparison with base model.

\begin{figure}[t]
  \includegraphics[width=1\linewidth]{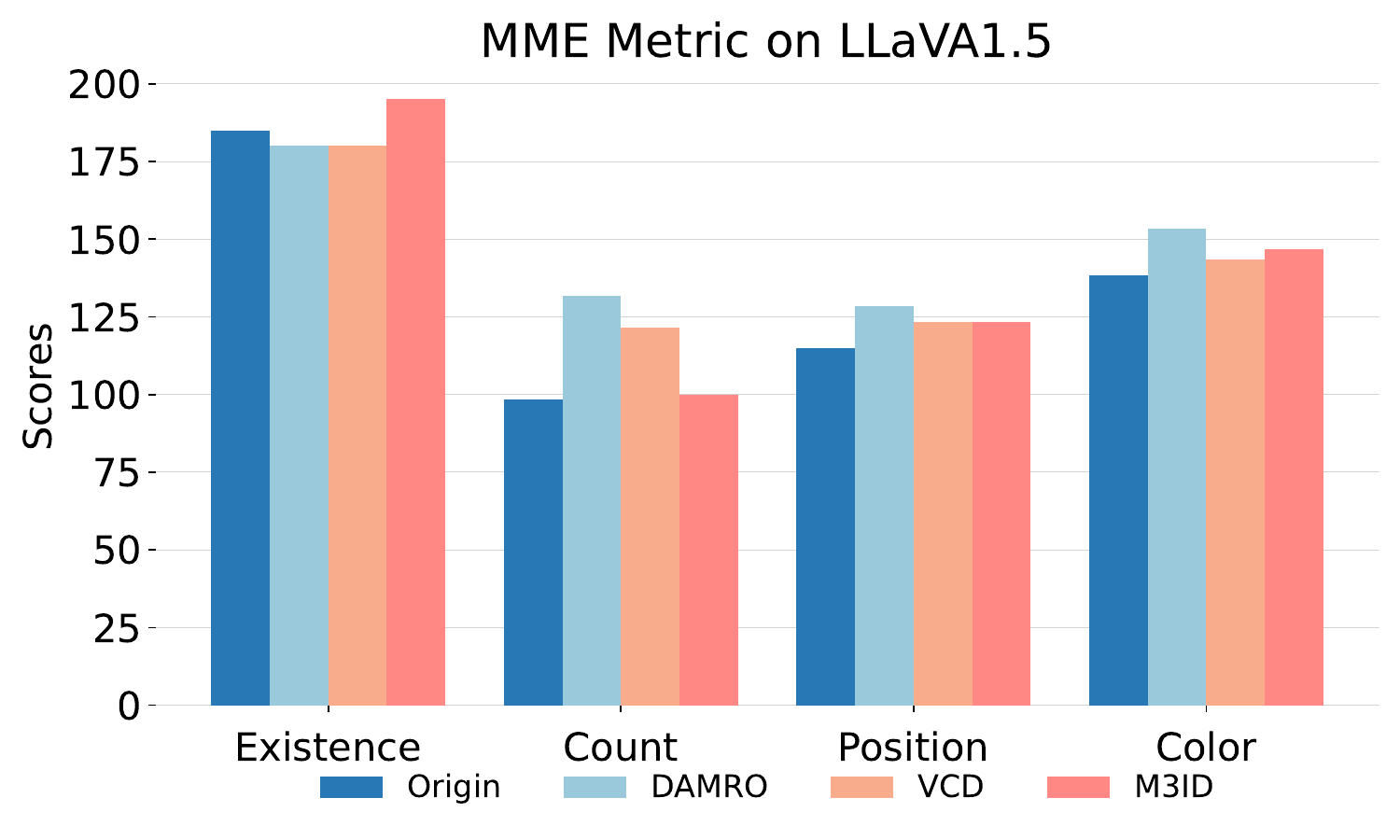}
  \hfill
  \includegraphics[width=1\linewidth]{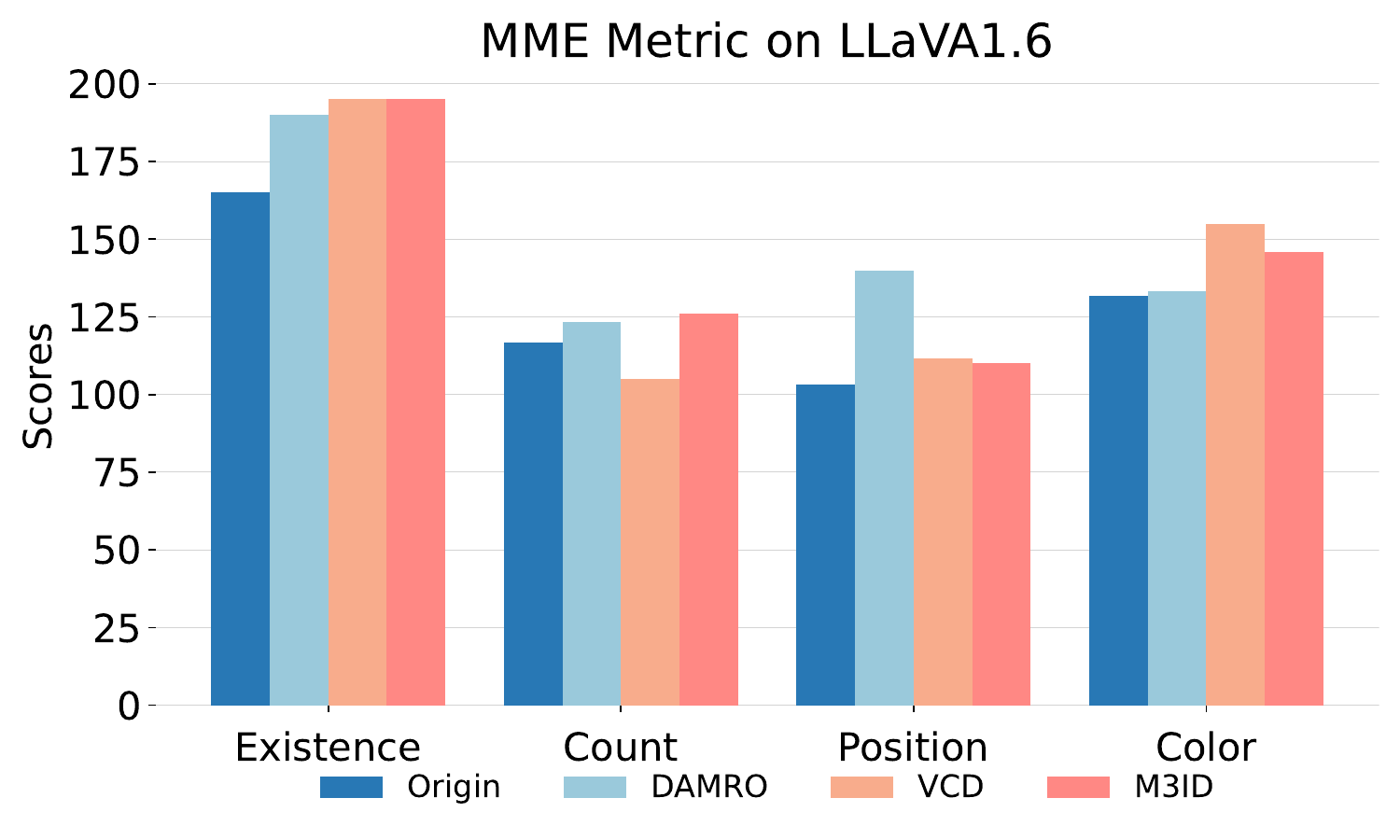}
   \hfill
  \includegraphics[width=1\linewidth]{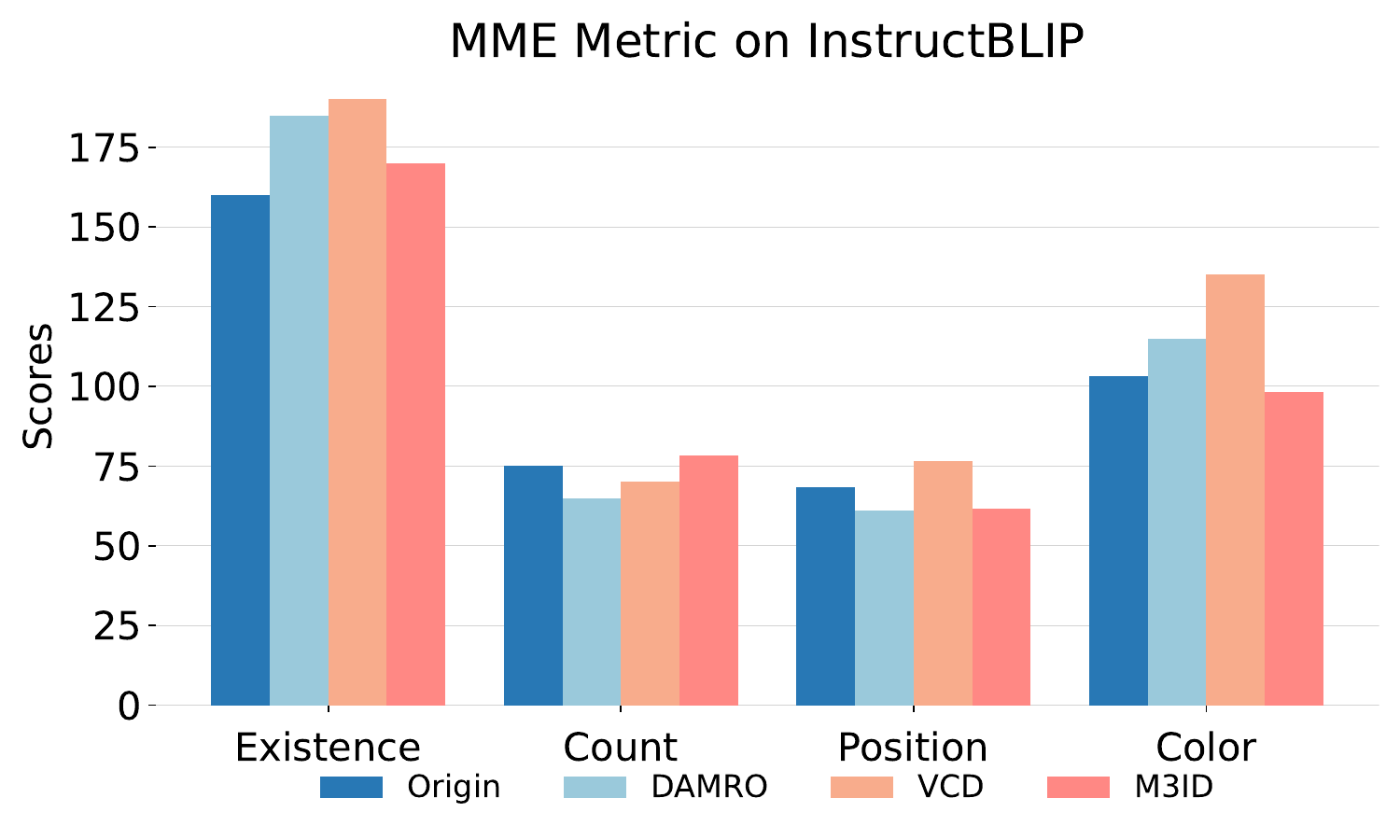}
   \hfill
  \caption {Results of MME.}
  \label{fig:mme}
\end{figure}

\paragraph{MME Hallucination Subset}

The Multimodal Large Language Model Evaluation (MME) \citep{fu2024mme} assesses LVLMs using a set of comprehensive metrics. Following the methodologies of \citet{yin2023woodpecker} and \citet{Leng_2024_CVPR}, we adopted "existence" and "count" from the MME benchmark as object-level evaluation metrics, and "color" and "position" as attribute-level evaluation metrics. The experimental results in Figure~\ref{fig:mme} demonstrate that our approach generally improves performance across three models, confirming its effectiveness. However, for InstructBLIP, metrics for count and position show a decline. We hypothesize that this is due to the unique structure of InstructBLIP, which relies on certain outlier tokens for spatial reasoning. Compared to the LLaVA series of foundation models, InstructBLIP has significantly weaker positional capabilities, possibly explaining the reduced effectiveness of our approach for this model. Experiment Details are shown in Appendix~\ref{sec:de_mme}.

\begin{table}[]
\begin{tabular}{cccc}
\hline
\textbf{Model} & \textbf{Method} & \textbf{A} & \textbf{D} \\ \hline
LLaVA-1.5            & Original        & 5.356             & 5.067                 \\
                    & \textbf{DAMRO}  & \textbf{6.611}    & \textbf{6.078}        \\ \hline
LLaVA-NeXT          & Original        & 6.456             & 6.332                 \\
                    & \textbf{DAMRO}  & \textbf{7.189}    & \textbf{6.656}        \\ \hline
InstructBLIP        & Original        & 5.833             & 5.400                 \\
                    & \textbf{DAMRO}  & \textbf{6.756}    & \textbf{5.967}        \\ \hline
\end{tabular}
\caption{Results of GPT4V-aided evaluation. A: accuracy, D: detailedness.}
\label{tab:gpt4v}
\end{table}

\paragraph{GPT4V-Aided Evaluation}
The GPT-4V-aided evaluation employs GPT-4V\footnote{\url{https://openai.com/index/gpt-4v-system-card/}} as an evaluator to compare the outputs of two LVLM assistants. GPT-4V assigns scores out of 10 based on two criteria: 1) accuracy, which measures how accurately each assistant describes the image, and 2) detailedness, which evaluates the richness of necessary details in the responses. 
We select LLaVA-QA90 \footnote{\url{https://github.com/haotian-liu/LLaVA/blob/main/playground/data/coco2014_val_gpt4_qa_30x3.jsonl}} 
for our tests on GPT-4V. The dataset consists of 30 images from COCO val2014, each paired with 3 questions to comprehensively evaluate the capabilities of LVLMs. Table \ref{tab:gpt4v} presents the overall scores of GPT-4V in terms of accuracy and detailedness, with detailed results provided in the appendix~\ref{sec:de_gpt4v}.

\section{Conclusions}
In this paper, we investigate the relationship between the attention maps of the visual encoder and the LLM decoder, and explore its impact on the mechanism of object hallucination in LVLMs. Based on our analysis of attention mechanism, we propose the Dive into Attention Mechanism to mitigate object hallucination (DAMRO) method. Our method demonstrates its effectiveness and generalizability on various models and benchmarks. Experiments show that our method effectively reduces hallucination issues in LVLMs across multiple domains, especially in fine-grained semantic hallucinations. Additionally, we hope our findings on Encoder-Decoder attention mechanism will inspire further research on LVLM foundation model structures.

\section*{Limitations}
Our method (DAMRO) is based on the relationship between the attention mechanisms of the visual encoder and the LLM decoder. It relies solely on empirical analysis and lacks further theoretical proof. Additionally, we have not conducted a detailed exploration of more complex projection modules in the visual encoder and LLM decoder (e.g. QFormer~\citep{10384565}). With the rapid development and continual refinement of LVLM models, whether our method remains applicable to future models poses a significant challenge.

\section*{Acknowledgements}
The work is partially supported by the National Nature Science Foundation of China (No. 62376199, 62076184, 62076182) and Shanghai Science and Technology Plan Project (No.21DZ1204800).


\appendix \label{sec:appendix}

\section{More Implementation Details} \label{sec:impl}

For the baselines M3ID and VCD, we employ the same direct sampling strategy as DAMRO. Throughout the entire experiment, our experimental hyperparameters remain consistent. The hyperparameters are listed in the table below:

\begin{table}[t]
\begin{tabular}{ll}
\hline
\textbf{Hyperparameters} & \textbf{Value}  \\ \hline
Forgetting Factor(POPE) $\gamma$           & 0.2  \\
Forgetting Factor(CHAIR, MME) $\gamma$           & 0.01  \\                   
       Threshold    & 0.9    \\ \hline
\end{tabular}
\caption{M3ID Hyperparameters Settings.}
\label{tab:gpt4v}
\end{table}

\begin{table}[t]
\begin{tabular}{ll}
\hline
\textbf{Hyperparameters} & \textbf{Value}  \\ \hline
Amplification Factor $\alpha$           & 1  \\
                   
Adaptive Plausibility Threshold    $\beta$     &       0.1   \\
               
Diffusion Noise Step        &  999  \\ \hline
\end{tabular}
\caption{VCD Hyperparameters Settings.}
\label{tab:gpt4v}
\end{table}

\section{Ablation Study } \label{sec:abla}

Considering that CHAIR can more precisely assess the generative capabilities of the model, and given that LLaVA-1.5 and LLaVA-NeXT have similar model structures, we choose to test the parameter sensitivity of DAMRO on LLaVA-1.5 and InstructBLIP using CHAIR. The following two parameter ablation experiments are based on this setup. As for how many visual tokens are enough, we conduct ablation experiments on LLaVA-1.5 using POPE, CHAIR and MME benchmarks.

\subsection{Effect of $\alpha$ in Visual Contrastive Decoding }

\begin{figure}[t]
\centering
  \includegraphics[width=1\linewidth]{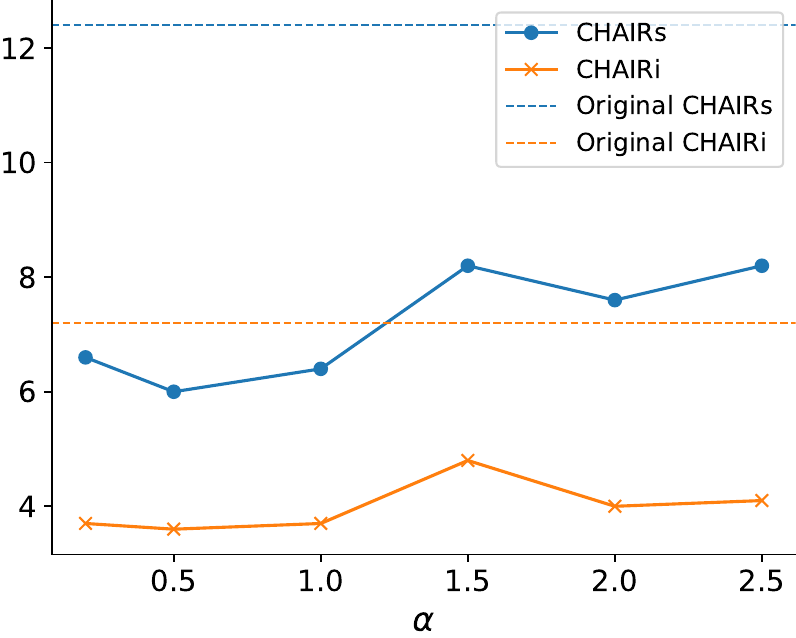}

  \caption {Ablation study of $\alpha$ in LLaVA-1.5, top k=10.}
  \label{fig:alpha1}
\end{figure}
\begin{figure}[t]
\centering
  \includegraphics[width=1\linewidth]{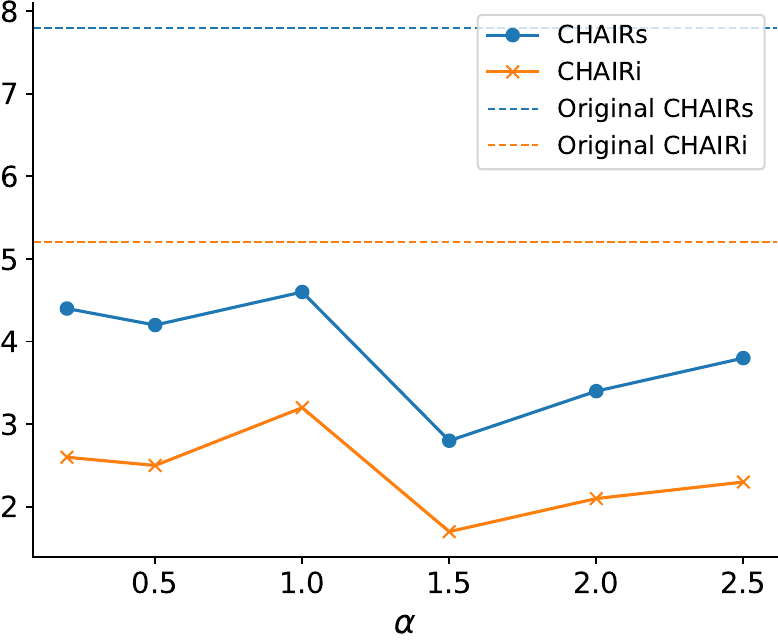}

  \caption {Ablation study of $\alpha$ in InstructBLIP, top k=4.}
  \label{fig:alpha2}
\end{figure}

The results of the experiments with LLaVA-1.5 and InstructBLIP are shown in Figure~\ref{fig:alpha1} and Figure~\ref{fig:alpha2}. It can be observed that when the value of $\alpha$ is too large or too small, the performance of the models deteriorates. $\alpha$ highlights the adjustment strength for outliers in our method, and the optimal adjustment strength varies for different models.

\subsection{ Effect of Outlier Token Number top $k$}

\begin{figure}[t]
\centering
  \includegraphics[width=1\linewidth]{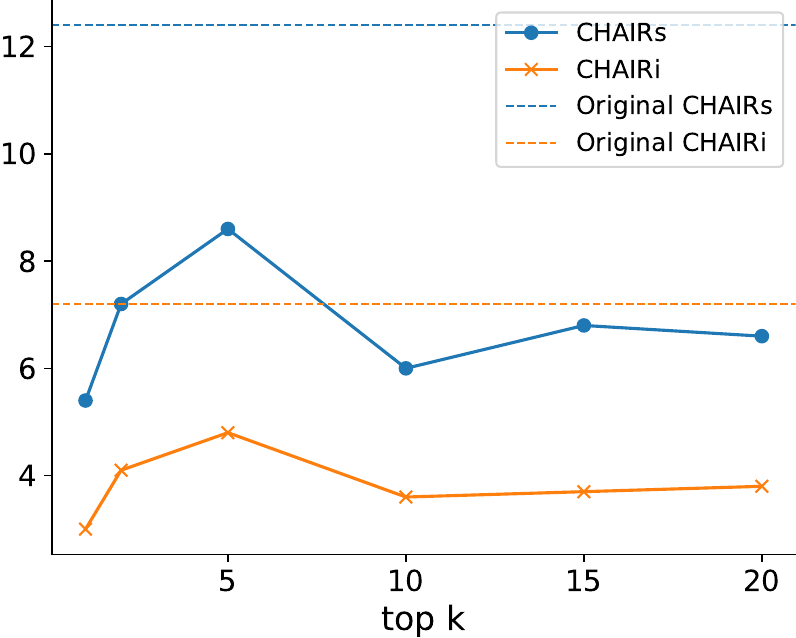}

  \caption {Ablation study of top $k$ in LLaVA-1.5, $\alpha$=0.5.}
  \label{fig:topk1}
\end{figure}
\begin{figure}[t]
\centering
  \includegraphics[width=1\linewidth]{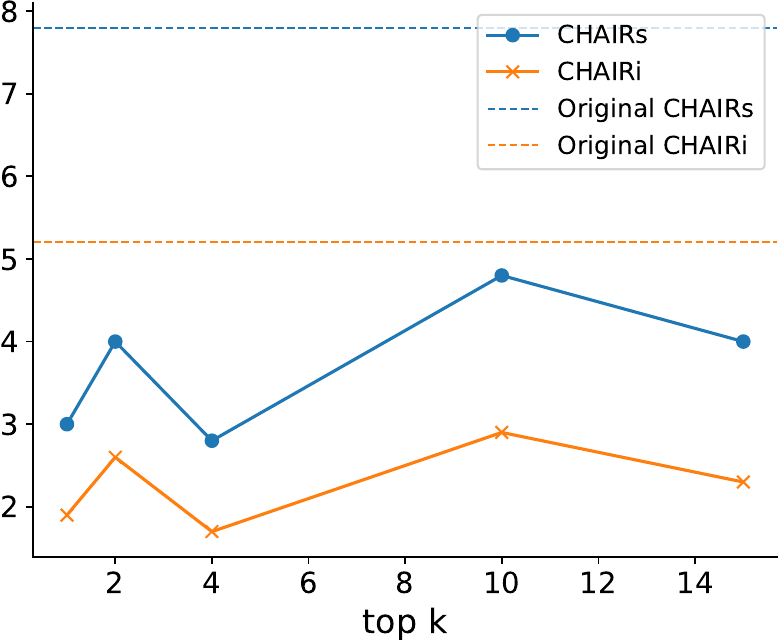}

  \caption {Ablation study of top $k$ in InstructBLIP, $\alpha$=1.5.}
  \label{fig:topk2}
\end{figure}

We use hyperparameters to define the number of outlier tokens, which vary across different visual encoders. Removing the top k outlier tokens aims to eliminate the redundant negative information they carry. However, this redundant information also contains a certain degree of global information, which can be beneficial for the results. Therefore, it is crucial to reasonably select the top k for our method. The results of the ablation experiments are shown in Figure~\ref{fig:topk1} and Figure~\ref{fig:topk2}.

\subsection{How Many Visual Tokens are Enough}\label{sec:token}

\begin{figure*}[]
\centering
  \includegraphics[width=1\linewidth]{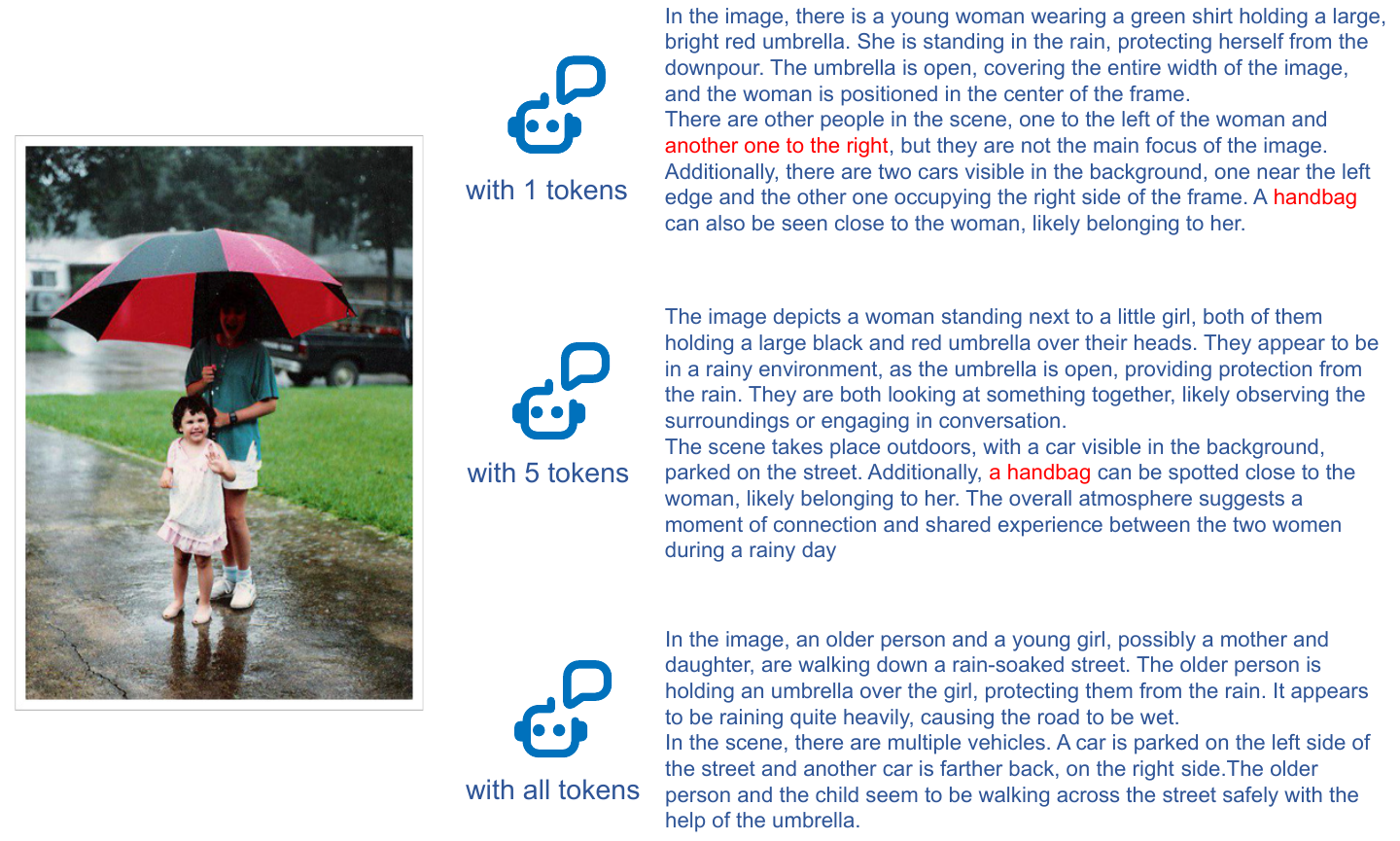}

  \caption {A case illustrates the generative ability of tokens. We use the prompt "Please describe this image in detail." to get answers from different. Hallucinated words are marked in red.}
  \label{fig:res}
\end{figure*}

\begin{figure*}[]
\centering
  \includegraphics[width=1\linewidth]{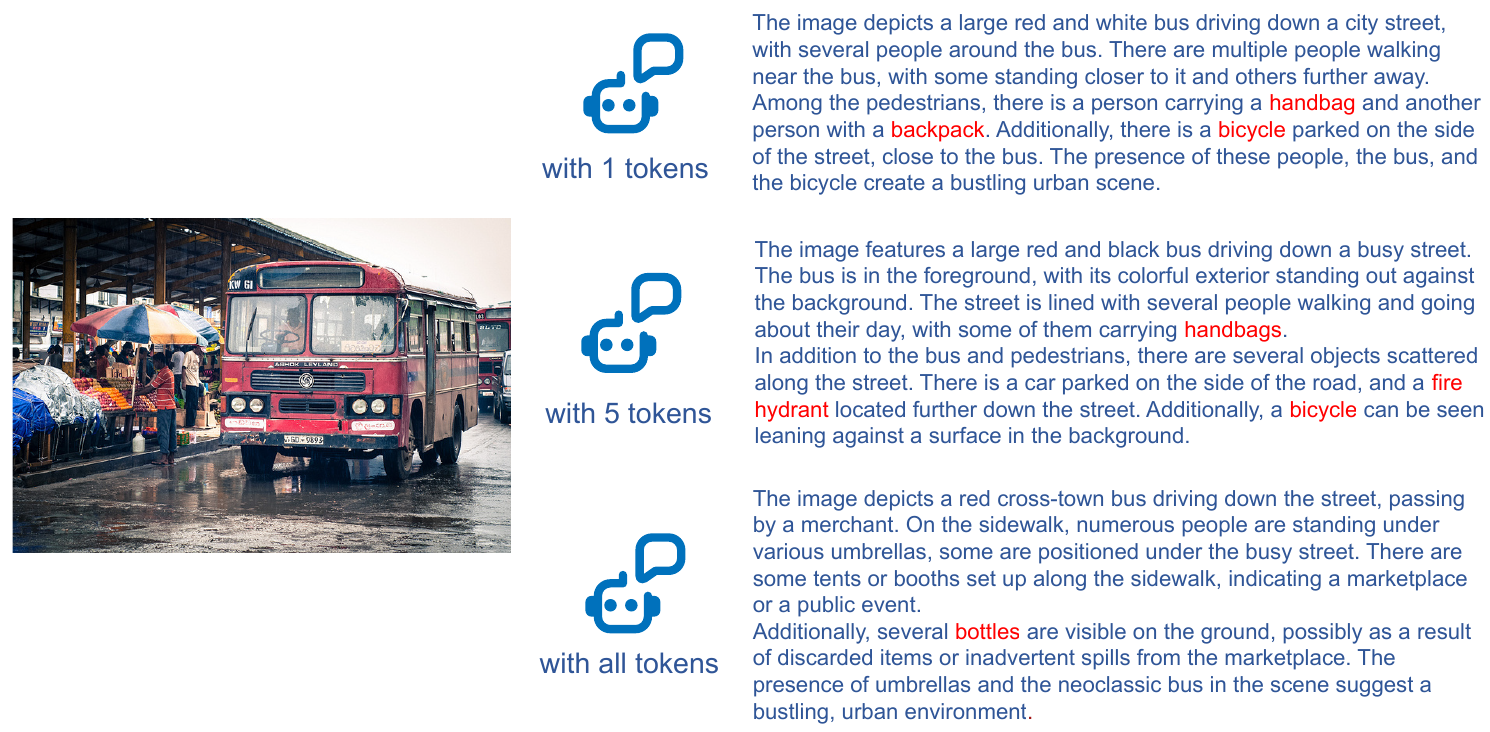}

  \caption {A case illustrates the generative ability of tokens. We use the prompt"Please describe this image in detail." to get the answers. Hallucinated words are marked in red.}
  \label{fig:ress}
\end{figure*}

We conduct experiments using LLaVA-1.5 on CHAIR, POPE(only on random split), and MME, and found that a small number of visual tokens, or even a single token, can contain the basic information of an entire image. POPE,CHAIR,MME results are shown in Table~\ref{tab:res1},Table~\ref{tab:res2} and Table~\ref{tab:res3} respectively. Additionally, we select some images and examples from these CHAIR experiments, as shown in Figure~\ref{fig:res} and Figure~\ref{fig:ress}. It is evident that a few tokens indeed contain a large amount of information. However, the error rate of this information is quite high, easily leading to the co-occurrence of related objects, which reflects the priors of the visual encoder.


An interesting phenomenon is that using only a small number of tokens, some metric results are actually better than using more tokens. We attribute this to the fact that the LLM's attention to visual tokens cannot accurately capture the information they contain. Therefore, this also provides an idea for better selection and acquisition of effective tokens in future LVLM models.

\begin{table}[H]
\begin{tabular}{ccccc}
\hline
             & \textbf{Precision} & \textbf{Recall} & \textbf{F1}    & \textbf{Accuracy} \\ \hline
top 1        & 89.93              & 70.87           & 79.27          & 81.47             \\
top 2        & 90.47              & 69.60            & 78.67          & 81.13             \\
top 5        & 93.29              & 64.93           & 76.57          & 80.13             \\
top 10       & 94.76              & 63.93           & 76.35          & 80.20              \\
top 100      & 95.50               & 66.47           & 78.38          & 81.67             \\ \hline
\textbf{all} & \textbf{92.32}     & \textbf{73.73}  & \textbf{81.97} & \textbf{83.80}     \\ \hline
\end{tabular}
\caption{POPE results with token numbers changed.}
\label{tab:res1}
\end{table}

\begin{table}[]
\centering
\begin{tabular}{cccc}
\hline
       & \textbf{CHAIRs $\downarrow$} & \textbf{CHAIRi  $\downarrow$} & \textbf{Recall $\uparrow$} \\ \hline
top1   & 58.6            & 18.4            & 61.4            \\
top2   & 53.6            & 17.0            & 61.0            \\
top5   & 57.8            & 15.1            & 67.0            \\
top10  & 50.6            & 14.4            & 60.5            \\
top100 & 57.8            & 15.1            & 67.0            \\ \hline
\textbf{all}    & \textbf{60.2}            & \textbf{16.8  }          &\textbf{ 68.1 }           \\ \hline
\end{tabular}
\caption{CHAIR results with token numbers changed.}
\label{tab:res2}
\end{table}

\begin{table*}[]
\centering
\begin{tabular}{cccccc}
\hline
             & \textbf{existence} & \textbf{count} & \textbf{position} & \textbf{color}  & \textbf{total} \\ \hline
top1         & 175.00             & 81.67          & 98.33             & 116.67          & 471.67         \\
top2         & 180.00             & 91.67          & 96.66             & 136.66          & 504.99         \\
top5         & 168.33             & 90.00          & 116.67            & 125.00          & 500.00         \\
top10        & 178.33             & 80.00          & 96.66             & 118.33          & 473.32         \\
top100       & 170.00             & 80.00          & 90.00             & 121.67          & 461.67         \\ \hline
\textbf{all} & \textbf{185.00}    & \textbf{98.30} & \textbf{115.00}   & \textbf{138.30} & \textbf{536.30 }        \\ \hline
\end{tabular}
\caption{MME results with token numbers changed.}
\label{tab:res3}
\end{table*}

\begin{table*}[]
\centering
\begin{tabular}{lllllll}
\hline

\textbf{Model}       & \textbf{Dataset }    & \textbf{Method}         & \textbf{Precision}       & \textbf{Recall    }      & \textbf{F1 }             & \textbf{Accuracy}        \\ \hline
LLaVA-1.5    & random      & Original       & 92.321          & 73.733          & 81.987          & 83.800          \\
            &             & \textbf{DAMRO} & \ul{94.557}    & \ul{81.067}    & \ul{87.294}    & \textbf{88.200} \\
            &             & VCD            & 91.886          & \textbf{83.8}   & \textbf{87.657} & \textbf{88.200} \\
            &             & M3ID           & \textbf{96.331} & 75.267          & 84.506          & \ul{86.200}    \\ \cline{2-7} 
            & popular     & Original       & \ul{89.700}    & 73.733          & 80.937          & 82.633          \\
            &             & \textbf{DAMRO} & 89.280          & \ul{81.067}    & \ul{84.976}    & \ul{85.667}    \\
            &             & VCD            & 87.231          & \textbf{83.800}   & \textbf{85.481} & \textbf{85.767} \\
            &             & M3ID           & \textbf{92.923} & 75.267          & 83.168          & 84.767          \\ \cline{2-7} 
            & adversarial & Original       & \ul{83.864}    & 73.800          & 78.511          & 79.800          \\
            &             & \textbf{DAMRO} & 82.677          & 81.133          & \textbf{81.898} & \textbf{82.067} \\
            &             & VCD            & 79.343          & \textbf{83.733} & \ul{81.479}    & 80.967          \\
            &             & M3ID           & \textbf{88.185} & 75.133          & 81.138          & \ul{82.533}    \\ \hline
LLaVA-NeXT    & random      & Original       & \ul{96.500}    & 75.600          & 84.785          & 86.433          \\
            &             & \textbf{DAMRO} & 94.749          & \textbf{85.400} & \textbf{89.832} & \textbf{90.333} \\
            &             & VCD            & 96.187          & \ul{82.400}    & \ul{88.760}     & \ul{89.567}    \\
            &             & M3ID           & \textbf{97.457} & 79.200          & 87.385          & 88.567          \\ \cline{2-7} 
            & popular     & Original       & 92.571          & 75.600          & 83.229          & 84.767          \\
            &             & \textbf{DAMRO} & 90.594          & \textbf{85.400} & \textbf{87.920}  & \textbf{88.267} \\
            &             & VCD            & 92.170          & \ul{82.400}    & \ul{87.010}    & \ul{87.700}    \\
            &             & M3ID           & \textbf{93.913} & 79.200          & 85.931          & 87.033          \\ \cline{2-7} 
            & adversarial & Original       & \ul{87.761}    & 75.533          & 81.189          & 82.500          \\
            &             & \textbf{DAMRO} & 84.720          & \textbf{85.400} & \textbf{85.059} & 85.000          \\
            &             & VCD            & 87.340          & \ul{ 82.400}    & 84.803          & \ul{85.233}    \\
            &             & M3ID           & \textbf{91.314} & 79.200          & \ul{84.827}    & \textbf{85.833} \\ \hline
InstrucBLIP & random      & Original       & 81.975          & 79.133          & 80.523          & 80.867          \\
            &             & \textbf{DAMRO} & 85.890          & \textbf{84.000} & \textbf{84.934} & \ul{85.100}    \\
            &             & VCD            & \ul{89.694}    & \ul{80.067}    & \ul{84.607}    & \textbf{85.433} \\
            &             & M3ID           & \textbf{93.451} & 70.400          & 80.304          & 82.733          \\ \cline{2-7} 
            & popular     & Original       & 79.112          & 79.067          & 79.093          & 79.100           \\
            &             & \textbf{DAMRO} & 80.089          & \textbf{83.667} & \textbf{81.839} & 81.433          \\
            &             & VCD            & \ul{83.907}    & \ul{79.600}    & \ul{81.697}    & \textbf{82.167} \\
            &             & M3ID           & \textbf{90.000} & 70.800          & 79.254          & \ul{81.467}    \\ \cline{2-7} 
            & adversarial & Original       & 74.829          & 80.067          & 77.359          & 76.567          \\
            &             & \textbf{DAMRO} & 76.010           & \textbf{84.067} & \textbf{79.835} & 78.767          \\
            &             & VCD            & \ul{81.052}    & \ul{80.133}    & \ul{79.59}     & \textbf{80.700}   \\
            &             & M3ID           & \textbf{88.314} & 70.533          & 78.428          & \ul{80.600}   \\  \hline
\end{tabular}

\caption{Detailed results of POPE on different sub-datasets.}
\label{tab:pope_all}
\end{table*}

\section{Detailed Results on POPE, MME and GPT4V-Aided Evaluation}
\subsection{POPE Details} \label{sec:de_pope}
The detailed results of POPE on different sub-datasets are shown in Table~\ref{tab:pope_all}.Our method achieved excellent results across different subsets.

\subsection{MME Details} \label{sec:de_mme}
The detailed results of MME  are shown in Table~\ref{tab:mme_all}

\begin{table*}[]
\centering
\begin{tabular}{ccccccc}
\hline
{\textbf{Model}} &{\textbf{Method}} & \multicolumn{2}{c}{\textbf{Object-level}} & \multicolumn{2}{c}{\textbf{Attribute-level}} & {\textbf{Total Scores}} \\ \cline{3-6}
                                &                                  & \textbf{Existence}    & \textbf{Count}    & \textbf{Position}      & \textbf{Color}      &                                        \\    
                                \hline
LLaVA-1.5                        & Original                         & \ul{ 185.00}          & 98.30             & 115.00                 & 138.30              & 536.60                                 \\
                                & VCD                              & \textbf{195.00}       & 100.00            & \ul{ 123.33}           & \ul{ 146.67}        & 565.00                                 \\
                                & M3ID                             & 180.00                & \ul{ 121.67}      & \ul{ 123.33}           & 143.33              & \ul{ 568.33}                           \\
                                & \textbf{DAMRO}                   & 180.00                & \textbf{131.67}   & \textbf{128.30}        & \textbf{153.30}     & \textbf{593.27}                        \\ 
                                  \noalign{\vskip 0.8ex}   \hline
LLaVA-NeXT                      & Original                         & 165.00                & 116.67            & 103.33                 & 131.66              & 516.66                                 \\
                                & VCD                              & \textbf{195.00}       & \textbf{126.00}   & 110.00                 & \ul{ 146.00}        & \ul{ 577.00}                           \\
                                & M3ID                             & \textbf{195.00}       & 105.00            & \ul{ 111.67}           & \textbf{155.00}     & 566.67                                 \\
                                & \textbf{DAMRO}                   & \ul{ 190.00}          & \ul{ 123.33}      & \textbf{140.00}        & 133.33              & \textbf{586.66}                        \\
                                  \noalign{\vskip 0.8ex}  \hline
InstructBLIP                    & Original                         & 160.00                & \ul{ 75.00}       & \ul{ 68.30}            & 103.3               & 406.60                                 \\
                                & VCD                              & 170.00                & \textbf{78.30}    & 61.67                  & 98.33               & 408.30                                 \\
                                & M3ID                             & \textbf{190.00}       & 70.00             & \textbf{76.67}         & \textbf{135.00}     & \textbf{471.67}                        \\
                                & \textbf{DAMRO}                   & \ul{185.00}          & 65.00             & 61.00                  & \ul{115.00}        & \ul{426.00}                           \\ 
                                  \noalign{\vskip 0.8ex}   \hline
\end{tabular}

\caption{Detailed results of MME.}
\label{tab:mme_all}
\end{table*}

\subsection{GPT4V-aided Evaluation Details} \label{sec:de_gpt4v}

\begin{figure*}[t]
  \includegraphics[width=1.0\linewidth]{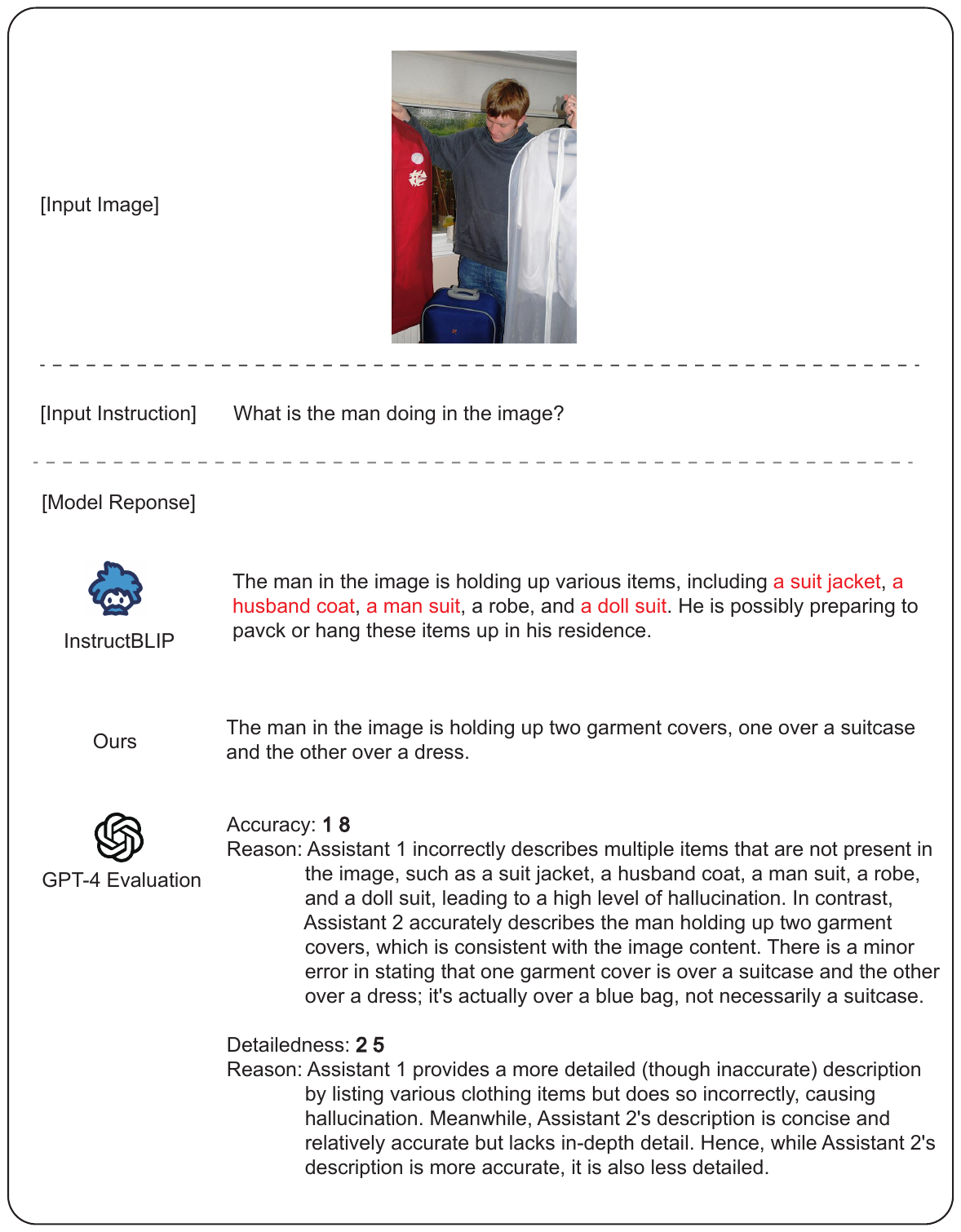}
  \caption {DAMRO's performance on reducing hallucinations on on InstructBLIP.}
  \label{fig:case1}
\end{figure*}

\begin{figure*}[t]
  \includegraphics[width=1.0\linewidth]{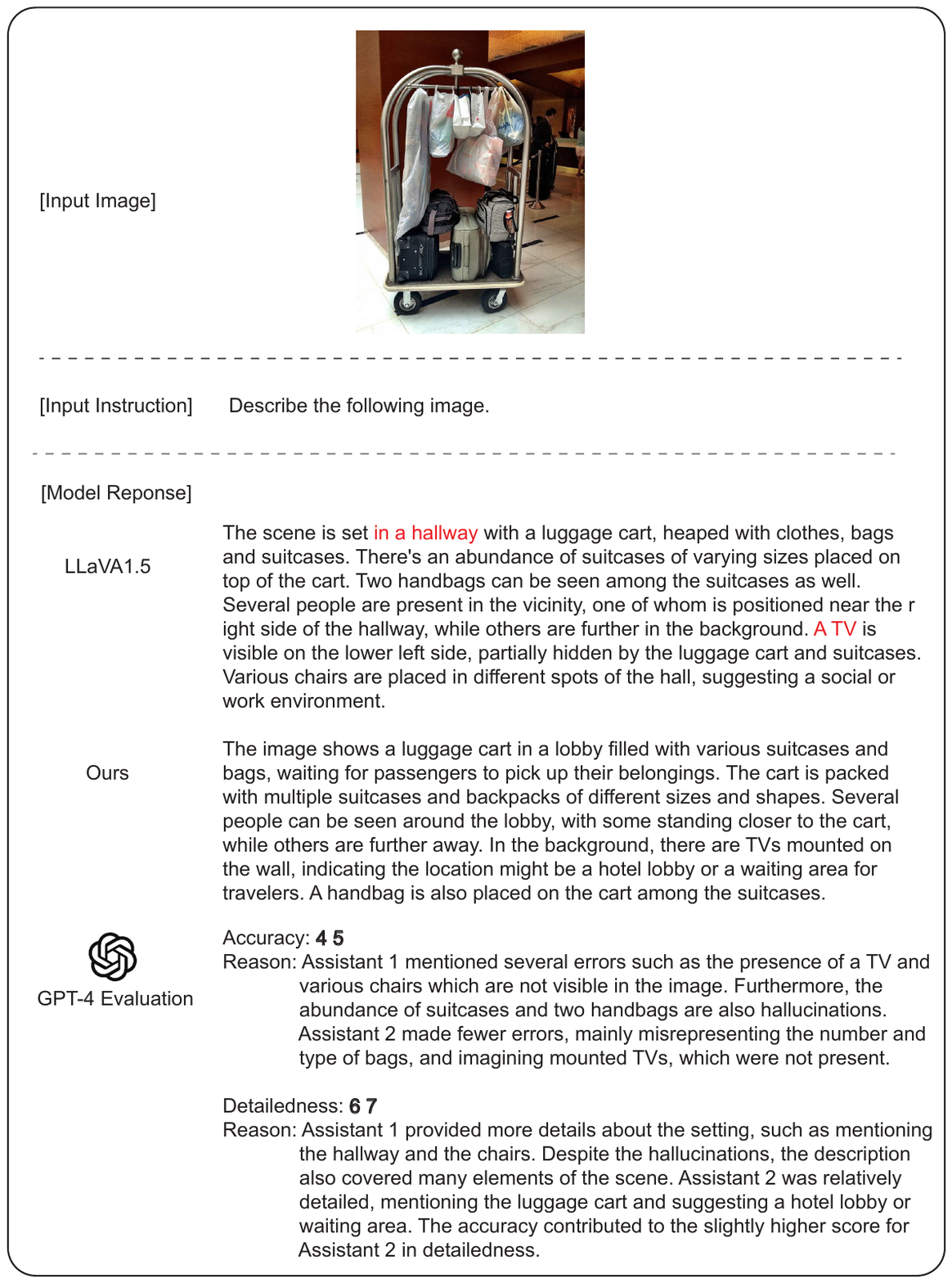}
  \caption {DAMRO's performance on reducing hallucinations on LLaVA-1.5-7b.}
  \label{fig:case2}
\end{figure*}

To evaluate open-ended generation, we utilize GPT-4V to assess the accuracy and detailedness of LVLMs’ responses. The specific configurations are detailed in Table~\ref{tab:confgpt}. Additionally, two illustrative evaluation cases are presented in Figure~\ref{fig:case1} and Figure~\ref{fig:case2}.

\begin{table*}[]
\begin{tabular}{l}
\hline
\textbf{GPT-4V(ision) Prompt}                                                                                                                                                                                                                                                                                                                                                                                                                                                                                                                                                                                               \\ \hline
\begin{tabular}[c]{@{}l@{}}You are required to score the performance of two AI assistants in describing a given image. You\\should pay extra attention to the hallucination, which refers to the part of descriptions that are\\inconsistent with the image content, such as claiming the existence of something not present in the\\image or describing incorrectly in terms of the counts, positions, or colors of objects in the image.\\Please rate the responses of the assistants on a scale of 1 to 10, where a higher score indicates\\better performance, according to the following criteria:\end{tabular} \\
\begin{tabular}[c]{@{}l@{}}1: Accuracy: whether the response is accurate with respect to the image content. Responses with\\fewer hallucinations should be given higher scores.\end{tabular}                                                                                                                                             \\
\begin{tabular}[c]{@{}l@{}}2: Detailedness: whether the response is rich in necessary details. Note that hallucinated descriptions\\should not count as necessary details.\end{tabular}                                                                                                                                                                                                                    \\
\begin{tabular}[c]{@{}l@{}}Please output the scores for each criterion, containing only two values indicating the scores for\\Assistant 1 and 2, respectively. The two scores are separated by a space. Following the scores, please\\provide an explanation of your evaluation, avoiding any potential bias and ensuring that the order\\in which the responses were presented does not affect your judgment.\end{tabular}                                                                         \\
{[}Assistant 1{]}                                                                                                                                                                                                                                                                          \\
\{\}                                                                                                                                                  \\
{[}End of Assistant 1{]}                                                                                                                                                                                             \\
       \\
{[}Assistant 2{]}                                                                                                                                                                                                                                                                   \\
\{\}                                                                                                                                                                                                                 \\
{[}End of Assistant 2{]}                                                                                                                                                                                                  \\
   \\
Output format:                                                                                                                                                                                                                     \\
Accuracy:                                                                                                                                                                                                                \\
Reason:                                                                                                                                                                                                                   \\
Detailedness:                                                                                                                                           \\
Reason:                                                                                                                                                                                                                   \\ \hline
\end{tabular}
\caption{The prompt used for GPT-4V(ision) evaluation.}
\label{tab:confgpt}
\end{table*}

\end{document}